\documentclass[iicol,sn-basic, pdflatex]{sn-jnl}

\usepackage{graphicx}%
\usepackage{multirow}%
\usepackage{amsmath,amssymb,amsfonts}%
\usepackage{amsthm}%
\usepackage{mathrsfs}%
\usepackage[title]{appendix}%
\usepackage{xcolor}%
\usepackage{textcomp}%
\usepackage{manyfoot}%
\usepackage{booktabs}%
\usepackage{algorithm}%
\usepackage{algorithmicx}%
\usepackage{algpseudocode}%
\usepackage{listings}%
\usepackage{amssymb}
\usepackage{stfloats}

\theoremstyle{thmstyleone}%
\theoremstyle{thmstyletwo}%

\theoremstyle{thmstylethree}%

\begin{document}

\title[Article Title]{Depth from Coupled Optical Differentiation}


\author[1]{\fnm{Junjie} \sur{Luo}}\email{luo330@purdue.edu}

\author[1]{\fnm{Yuxuan} \sur{Liu}}\email{liu3910@purdue.edu}

\author*[2]{\fnm{Emma} \sur{Alexander}}\email{ealexander@northwestern.edu}

\author*[1]{\fnm{Qi} \sur{Guo}}\email{qiguo@purdue.edu}

\affil*[1]{\orgdiv{Elmore Family School of Electrical and Computer Engineering}, \orgname{Purdue University}, \orgaddress{\street{501 Northwestern Ave}, \city{West Lafayette}, \postcode{47906}, \state{Indiana}, \country{USA}}}

\affil[2]{\orgdiv{Department of Computer Science}, \orgname{McCormick School of Engineering and Applied Science, Northwestern University}, \orgaddress{\street{2233 Tech Drive}, \city{Evanston}, \postcode{60208}, \state{Illinois}, \country{USA}}}


\abstract{We propose depth from coupled optical differentiation, a low-computation passive-lighting 3D sensing mechanism. It is based on our discovery that per-pixel object distance can be rigorously determined by a coupled pair of optical derivatives of a defocused image using a simple, closed-form relationship. Unlike previous depth-from-defocus (DfD) methods that leverage spatial derivatives of the image to estimate scene depths, the proposed mechanism's use of only optical derivatives makes it significantly more robust to noise. Furthermore, unlike many previous DfD algorithms with requirements on aperture code, this relationship is proved to be universal to a broad range of aperture codes. \\

We build the first 3D sensor based on depth from coupled optical differentiation. Its optical assembly includes a deformable lens and a motorized iris, which enables dynamic adjustments to the optical power and aperture radius. The sensor captures two pairs of images: one pair with a differential change of optical power and the other with a differential change of aperture scale. From the four images, a depth and confidence map can be generated with only 36 floating point operations per output pixel (FLOPOP), more than ten times lower than the previous lowest passive-lighting depth sensing solution to our knowledge. Additionally, the depth map generated by the proposed sensor demonstrates more than twice the working range of previous DfD methods while using significantly lower computation. 









}




\maketitle

\pagebreak
\section{Introduction}

The capability to perceive object depths at very low power consumption and without using time-resolved or space-resolved illumination has been prevalent in nature. Jumping spiders, praying mantis, etc., have demonstrated such passive-lighting depth sensing capabilities in their visual systems~\citep{land2012animal}. However, it has been extremely challenging for humans to embed a passive-lighting depth sensor in miniature artificial systems, such as micro-robots~\citep{wood2013flight}, microsensors~\citep{park2012micro}, wearable or edible devices~\citep{perez2015detection}, etc. A major reason is that passive-lighting depth sensors typically require sophisticated computational operations to extract depth information from the raw measurements.

In recent years, a series of works has made remarkable breakthroughs towards low-computation, passive-lighting depth sensing using depth from defocus (DfD) as the cue to extract 3D information from defocused images~\citep{alexander2018focal, guo2017focal, guo2019compact}. They report several depth sensors that reconstruct sparse, per-pixel depth maps with as low as 600 floating point operations per output pixel (FLOPOP). As a reference, an efficient stereo algorithm that achieves similar depth estimation accuracy costs around 7,000 FLOPOP~\citep{rotheneder2018performance}. The reduction in computational cost is possible because these sensors transform a portion of the signal processing to be performed optically during the image formation process. 
However, these depth sensors demonstrate a small working range around the depth of field of the image, e.g., 10-20 cm with 5\%-10\% relative depth error~\citep{guo2019compact}. This is because the depth sensing algorithms rely on spatial derivatives of the images to calculate object depths, which is fundamentally challenging to measure accurately when the scene is far from the depth of field and the defocus blur smoothes out the textures in captured images. 

\begin{figure}
    \centering
    \includegraphics[width=\linewidth]{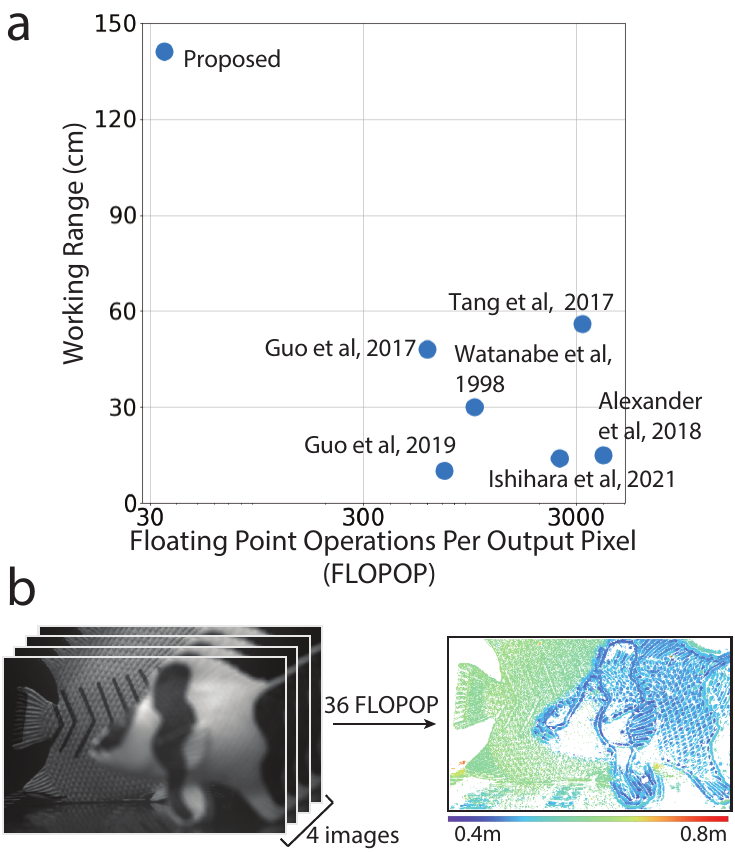}
    \caption{(a) Technological advantages of the proposed method. We plot the computational complexity, measured in floating point operations per output pixel (FLOPOP), and the working range of a series of efficient monocular, passive-lighting depth sensors. The proposed solution achieves a significantly lower computational complexity and longer working range compared to the previous best. (b) System diagram. The proposed depth sensor captures four images of a fixed scene with different optical settings and produces a sparse depth map with only 36 FLOPOP. }
    \label{fig:teaser}
\end{figure}

This work presents a new DfD-based depth sensor, which reports a more than ten times reduction of computational complexity and a more than two times increase in the working range compared to previous work~\citep{guo2019compact} (Fig.~\ref{fig:teaser}a). This specialized monocular sensor can capture images $I$ with dynamically controlled optical power $\rho$ and aperture radius $\mathnormal{A}$ using a deformable lens and a motorized iris to estimate image derivatives with respect to these two optical parameters, i.e., $I_\rho$ and $I_\mathnormal{A}$, via finite difference. A depth map can be estimated via a per-pixel calculation using the two image derivatives:
\begin{align}
    Z = \frac{a}{b - I_\rho / I_\mathnormal{A}},
    \label{eq:1}
\end{align}
where the parameters $a, b$ are pre-calibrated constants determined by the optical setup. An overview of the system and a sample depth map is shown in Fig.~\ref{fig:teaser}b.

The proposed depth sensor is based on a new 3D-sensing theory, \textit{depth from coupled optical differentiation}. The theory mathematically shows that depth can be estimated at every pixel using image derivatives with respect to the two optical parameters, i.e., $I_\rho$ and $I_\mathnormal{A}$ in Eq.~\ref{eq:1}. The theory proves that the correctness of Eq.~\ref{eq:1} is invariant to the scene appearance under the thin lens assumption. 
Compared with the previous depth from differential defocus (DfDD) theory~\citep{alexander2019theory}, which is restricted to only Gaussian point spread functions (PSFs), ours is invariant to differentiable PSF shapes. We also show that the proposed theory's signal-to-noise ratio (SNR) is significantly higher than that of DfDD, which explains why the proposed sensor can achieve a much longer working range and more accurate depth estimation than DfDD sensors. 

The contribution of the paper can be summarized as follows:
\begin{itemize}
    \item A mathematically rigorous depth-sensing theory named depth from coupled optical differentiation that is texture invariant and PSF-invariant. The theory suggests a new computationally efficient mechanism for 3D sensing.
    \item A comprehensive simulation analysis of depth from differential defocus, which explains the advantage in working range of depth from coupled optical differentiation and studies the effects of different optical and computational parameters. 
    \item A low-computation, monocular, and passive-lighting depth sensor with a data processing cost of only 36 FLOPOP, more than ten times lower than the most efficient depth sensing algorithms before. It also achieves a working range more than twice of the previous efficient DfD methods under the same optical setting.
\end{itemize}

\section{Related Work}

The most widely adopted 3D sensing methods based on optical wavelength electromagnetic signals can be classified into the following categories: time-of-flight, structured light, learning-based 2D-to-3D lifting, stereo, and depth from defocus (DfD). This section will briefly discuss the advantages and disadvantages of methods in each category and then review DfD-based approaches where the proposed method belongs. 

Time-of-flight~\citep{horaud2016overview} leverage time-resolved illumination to measure object distances by timing the round trip of the light signal from the emitter to the receiver. Structured light~\citep{zhang2018high, mirdehghan2018optimal, chen2020auto} utilizes space-resolved illumination to match key points of the scene from different viewing perspectives for triangulation. These active-illumination solutions provide a higher depth accuracy and a physically accurate dense depth map in a more controlled environment~\citep{koschan1997dense}. Meanwhile, they also have a higher hardware complexity and power consumption due to the required illumination and are susceptible to background noise and multipath interference of complex scene structures~\citep{foix2011lock, supreeth2017epipo, guo2018tackling}. 

Stereo estimates the disparities of corresponding key points in at least two images captured from different perspectives and calculates depth from the disparity via triangulation. As the depth estimation error is inverse-proportional to the baseline distance between the cameras of different perspectives~\citep{ding2011stereo}, stereo cameras typically use a baseline distance at least several times longer than the camera's aperture diameter for sufficient depth prediction accuracy~\citep{fan2020computer}. This often results in large disparities between corresponding key points, and sophisticated disparity matching algorithms, including learning-based~\citep{luo2016efficient} and non-learning-based~\citep{ploumpis2015stereo}, have to be used to detect the correspondences robustly. Meanwhile, people have also proposed micro-baseline stereo solutions and shown depth can be extracted with a relatively low computation but with a high error~\citep{farid1998range, joshi2014micro, wadhwa2018synthetic}. 

DfD is closely related to stereo as it is also based on triangulation, whereas the baseline distance of DfD is defined by the aperture diameter of the camera. Schechner and Kiryati show that DfD is preferred over stereo when the baseline distance is small~\citep{schechner2000depth}. This is because DfD enables a higher signal-to-noise ratio in its images by using a larger equivalent aperture and also ``allows much more pixels in the image to contribute to depth estimation"~\citep{schechner2000depth} than stereo. 

\begin{table*}[]
\hspace{-1in}
\caption{Comparison of low-computation depth from defocus (DfD) solutions. 
}
\footnotesize
\resizebox{\textwidth}{!} {
\begin{tabular}{|l|l|l|l|c|c|c|c|c|}
\hline
\multicolumn{1}{|c|}{\begin{tabular}[c]{@{}c@{}} Method \\ \end{tabular}}
& \multicolumn{1}{c|}{\begin{tabular}[c]{@{}c@{}} Venue \\ \end{tabular}}

& \multicolumn{1}{c|}{\begin{tabular}[c]{@{}c@{}}Optical \\ Mechanism\end{tabular}} 

& \multicolumn{1}{c|}{\begin{tabular}[c]{@{}c@{}}Processing \\ Algorithm\end{tabular}} 

& \multicolumn{1}{c|}{\begin{tabular}[c]{@{}c@{}c@{}} \#I$^1$ \end{tabular}} 

& \multicolumn{1}{c|}{\begin{tabular}[c]{@{}c@{}} FLOP-\\OP$^2$\end{tabular}}

& \multicolumn{1}{c|}{\begin{tabular}[c]{@{}c@{}c@{}}RoA$^{3}$ \\(cm)\end{tabular}}  

& \multicolumn{1}{c|}{\begin{tabular}[c]{@{}c@{}c@{}c@{}c@{}}Depth \\ Error$^4$ \\ (cm) \end{tabular}} 

& \multicolumn{1}{c|}{\begin{tabular}[c]{@{}c@{}c@{}}RF$^5$ \\ (pixel$^2$)\end{tabular}} \\ \hline

Proposed & \begin{tabular}[c]{@{}l@{}}
\end{tabular}
& \begin{tabular}[c]{@{}l@{}l@{}}Deformable \\lens + \\ dynamic \\aperture\end{tabular} & \begin{tabular}[c]{@{}l@{}l@{}}Coupled \\ optical \\ differentiation\end{tabular} & 4 & 36 & \multicolumn{1}{c|}{45 - 186} & \multicolumn{1}{c|}{4.4$^\blacklozenge$} & \begin{tabular}[c]{@{}c@{}}5$\times$5\end{tabular}\\ \hline

\begin{tabular}[c]{@{}l@{}} {\cite{ishihara2021depth}} \end{tabular} & \begin{tabular}[c]{@{}l@{}l@{}}JOSA \\ 2021
\end{tabular}
& \begin{tabular}[c]{@{}l@{}l@{}l@{}}Hyper-\\spectral \\ sensitive \\ pixel\end{tabular} & \begin{tabular}[c]{@{}l@{}l@{}}Spectral \\ differential \\ defocus\end{tabular}&6&2.5e3&\multicolumn{1}{c|}{90 - 104}&\multicolumn{1}{c|}{1.9$^\diamond$}& \begin{tabular}[c]{@{}c@{}}9$\times$9 \end{tabular}\\ \hline

\begin{tabular}[c]{@{}l@{}}{\cite{guo2019compact}}\end{tabular} & \begin{tabular}[c]{@{}l@{}l@{}}PNAS \\ 2019
\end{tabular}
& \begin{tabular}[c]{@{}l@{}l@{}}Multi-\\functional \\Metasurface \end{tabular} & \begin{tabular}[c]{@{}l@{}}Differential \\ defocus\end{tabular} & 2 & 7e2 & \multicolumn{1}{c|}{30 - 40} & \multicolumn{1}{c|}{0.3$^\blacklozenge$} & \begin{tabular}[c]{@{}c@{}}25$\times$25 \end{tabular}\\ \hline

\begin{tabular}[c]{@{}l@{}} {\cite{alexander2018focal}} \end{tabular} & \begin{tabular}[c]{@{}l@{}l@{}}IJCV \\ 2018
\end{tabular}
& \begin{tabular}[c]{@{}l@{}l@{}}Camera \\ or object\\ motion\end{tabular} & \begin{tabular}[c]{@{}l@{}}Differential \\ defocus\end{tabular} & 3$^*$ & 4e3 & \multicolumn{1}{c|}{50 - 65$^\blacktriangle$} & \multicolumn{1}{c|}{ 5.7$^\blacklozenge$} & \begin{tabular}[c]{@{}c@{}}71$\times$71 \end{tabular}\\ \hline

\begin{tabular}[c]{@{}l@{}}{\cite{guo2017focal}}$^\dagger$\end{tabular}& \begin{tabular}[c]{@{}l@{}l@{}}ICCV \\ 2017
\end{tabular}
& \begin{tabular}[c]{@{}l@{}l@{}}Liquid \\ deformable \\ lens \end{tabular} & \begin{tabular}[c]{@{}l@{}}Differential \\ defocus\end{tabular} & 3 & 6e2 & \multicolumn{1}{c|}{68 - 115} & \multicolumn{1}{c|}{6.0$^\blacklozenge$} & \begin{tabular}[c]{@{}c@{}}20$\times$20 \end{tabular}\\ \hline

\begin{tabular}[c]{@{}l@{}} {\cite{tang2017depth}}$^\ddagger$ \end{tabular} & \begin{tabular}[c]{@{}l@{}l@{}}CVPR \\ 2017
\end{tabular}
& \begin{tabular}[c]{@{}l@{}l@{}}Focus \\ setting\end{tabular} & \begin{tabular}[c]{@{}l@{}l@{}}Defocus \\ equalization \\ filter\end{tabular} & 2 & 3.2e3 & \multicolumn{1}{c|}{75 - 131} & \multicolumn{1}{c|}{4.6$^\blacklozenge$} & \begin{tabular}[c]{@{}c@{}}5$\times$5 \end{tabular}\\ \hline

\begin{tabular}[c]{@{}l@{}} {\cite{zhou2011coded}} \end{tabular} & \begin{tabular}[c]{@{}l@{}}IJCV \\2011
\end{tabular}
& \begin{tabular}[c]{@{}l@{}l@{}}Coded \\ aperture\end{tabular} & \begin{tabular}[c]{@{}l@{}l@{}}Deblurring \\ and \\ reblurring\end{tabular}  & 2 & 1e3 & \multicolumn{3}{c|}{\begin{tabular}[c]{@{}l@{}l@{}}No quantitative analysis \\for real data\end{tabular} } \\ \hline

\begin{tabular}[c]{@{}l@{}}  {\cite{watanabe1998rational}} \end{tabular}& \begin{tabular}[c]{@{}l@{}}IJCV \\ 1998
\end{tabular}
& \begin{tabular}[c]{@{}l@{}l@{}}Focus \\ setting\end{tabular}  & \begin{tabular}[c]{@{}l@{}}Rational \\ operator\end{tabular} & 2$^*$ & 1e3 & \multicolumn{1}{c|}{55 - 85$^\blacktriangle$}  & \multicolumn{1}{c|}{0.42$^\diamond$}  & \begin{tabular}[c]{@{}c@{}}5$\times$5 \end{tabular}\\ \hline


\end{tabular}
} 

\footnotesize

$^1$ The number of differently defocused monochrome images required to generate a depth map. Numbers with $*$ indicate that multiple frames were reported to be averaged to form one defocused image to suppress noise during the inference. \\
$^2$ The computational cost of each method. We provide an educated estimate of each method in floating point operations per output pixel (FLOPOP). As a reference, the computation of an efficient stereo algorithm is around 7,000 FLOPOP~\citep{guo2019compact}. \\
$^3$ The \textit{region of accuracy (RoA)} is defined as the closest and farthest object distance where the average depth error is $<10\%$ of the true depth. We also define the \textit{working range} as the length of RoA. For numbers with $\blacktriangle$, the RoA cannot be directly read from the results in the paper, and we provide an educated estimate of the RoA according to our definition. \\
$^4$ Overall depth errors within the RoA. Markers $\blacklozenge$ and $\diamond$ indicate the numbers are MAEs and RMSEs, respectively.  \\
$^5$ The receptive field (RF) indicates the pixel areas in the measured image used to predict one depth value. \\ 
$\dagger$ Numbers are reported from our re-implementation. Both the proposed method and~\cite{guo2017focal} can vary the optical powers. Here we only report the numbers with a fixed optical power.\\
$^\ddagger$ Only the local stage is considered because we evaluate sparse outputs, and the global stage is computationally expensive primarily due to densification. Numbers are reported from our re-implementation.




\label{tab:comparison}
\end{table*}

\subsection{Depth from Defocus}

Depth from defocus (DfD) algorithms use the defocus blur in images as a cue to estimate the depth map. 
Theoretically, DfD algorithms require capturing at least two images $I_i, i=1,\cdots, N, N\ge2$ of a static scene with different defocus blur to predict the depth map without ambiguity~\citep{szeliski2022computer}. Although people have demonstrated single-image DfD using priors such as natural image statistics~\citep{levin2007image}, this section will discuss DfD methods using more than one defocused image. 

Consider two images of a front-parallel object $I_1(x,y), I_2(x,y)$, each with a different defocus blur. Mathematically, the Fourier spectrum of the images, $\mathcal{F}(I_i(x,y)), i=1,2$, are proportional to each other, with the ratio being invariant to the scene texture and only related to the object depth, $Z$~\citep{guo2022efficient}:
\begin{align}
     Z = \text{DfD}\left(\frac{\mathcal{F}(I_1(x,y))}{\mathcal{F}(I_2(x,y))}\right),
    \label{eq:dfd}
\end{align}
where $\text{DfD}()$ is a mapping between the ratio of the spectrums and the depth $Z$. This simple, non-iterative relation can be well-generalized to objects with varying depths by approximating each patch of the object to be front-parallel~\citep{guo2019compact}. Most existing DfD algorithms are sophisticated variants of Eq.~\ref{eq:dfd} to be robust to image noise and artifacts~\citep{watanabe1998rational, tang2017depth, subbarao1994depth, zhou2011coded, farid1998range}. This includes using specially designed filters to attenuate noise in the image patches~\citep{watanabe1998rational, tang2017depth, alexander2019theory}, parametric priors~\citep{subbarao1994depth}, engineered aperture code~\citep{zhou2011coded, farid1998range}, differential defocus~\citep{alexander2019theory}, etc. Similar to Eq.~\ref{eq:dfd}, these algorithms typically have non-iterative computation, some even with closed-form solutions, and thus can be implemented with low computational complexity~\citep{guo2019compact}.

There is a complementary family of DfD algorithms that leverages deep neural networks to learn to generate depth maps from the defocused images~\citep{wu2019phasecam3d, chang2019deep, tan20213d, gur2019single} using data. These methods implicitly learn the mapping from the defocus blur to depth instead of using the explicit DfD cue in Eq.~\ref{eq:dfd}. They typically have a much higher computational complexity, e.g. 300,000 FLOPs per pixel~\citep{wu2019phasecam3d} but can directly output dense, well-refined depth maps. However, these methods typically do not have the option to generate a less-refined depth map with a lower computation. Thus, they are more suitable for applications where computational power is not a constraint.

The first DfD algorithm was introduced decades ago~\citep{pentland1987new}, but DfD prototypes with real-time and high-quality depth sensing capabilities only appeared in recent years~\citep{guo2017focal, guo2019compact}. This is because a DfD sensor requires fast-response, high-optical-performance, dynamic optical devices to capture the differently defocused images as required by the algorithm. Such devices have been accessible recently thanks to the maturation of various optical and nanophotonic technologies. People have demonstrated DfD sensor prototypes with fast oscillation deformable lenses~\citep{guo2017focal, sheinin2019depth}, multifunctional metasurfaces~\citep{guo2019compact}, diffractive-optical elements~\citep{wu2019phasecam3d}, hyperspectral sensitivity pixels~\citep{ishihara2019depth, ishihara2021depth}, color-coded apertures~\citep{mishima2019physical}, etc. For example, Guo et al. demonstrate a single-shot DfD prototype that can generate depth maps in real-time at 100 frames per second. The prototype consists of a multifunctional metasurface that forms two defocused images with different optical powers side-by-side on a photosensor simultaneously~\citep{guo2019compact}. A detailed comparison between different DfD systems is listed in Table~\ref{tab:comparison}.


As shown in Table~\ref{tab:comparison}, DfD sensors almost universally demonstrate small working ranges, typically shorter than 60 cm for a fixed optical configuration~\citep{guo2017focal, guo2019compact, tang2017depth}. This is because most previous DfD algorithms need to use derivatives filters, effectively highpass filters, to extract image intensity variation as the signal for DfD, which inevitably magnifies the image noise~\citep{alexander2018focal, guo2017focal, guo2019compact, subbarao1994depth, watanabe1998rational}. The signal-to-noise ratio becomes low when the object is out of the depth of field because the image intensity variation becomes less significant due to defocus blur compared to the noise. This poses a natural constraint on the working ranges of DfD methods. To overcome it, we must develop a DfD algorithm not based on spatial derivatives of the captured images.

\section{Theory}

\begin{figure*}
    \vspace{0.4in}
    \hspace{-0.2in}
    \includegraphics[width=1.0\textwidth]{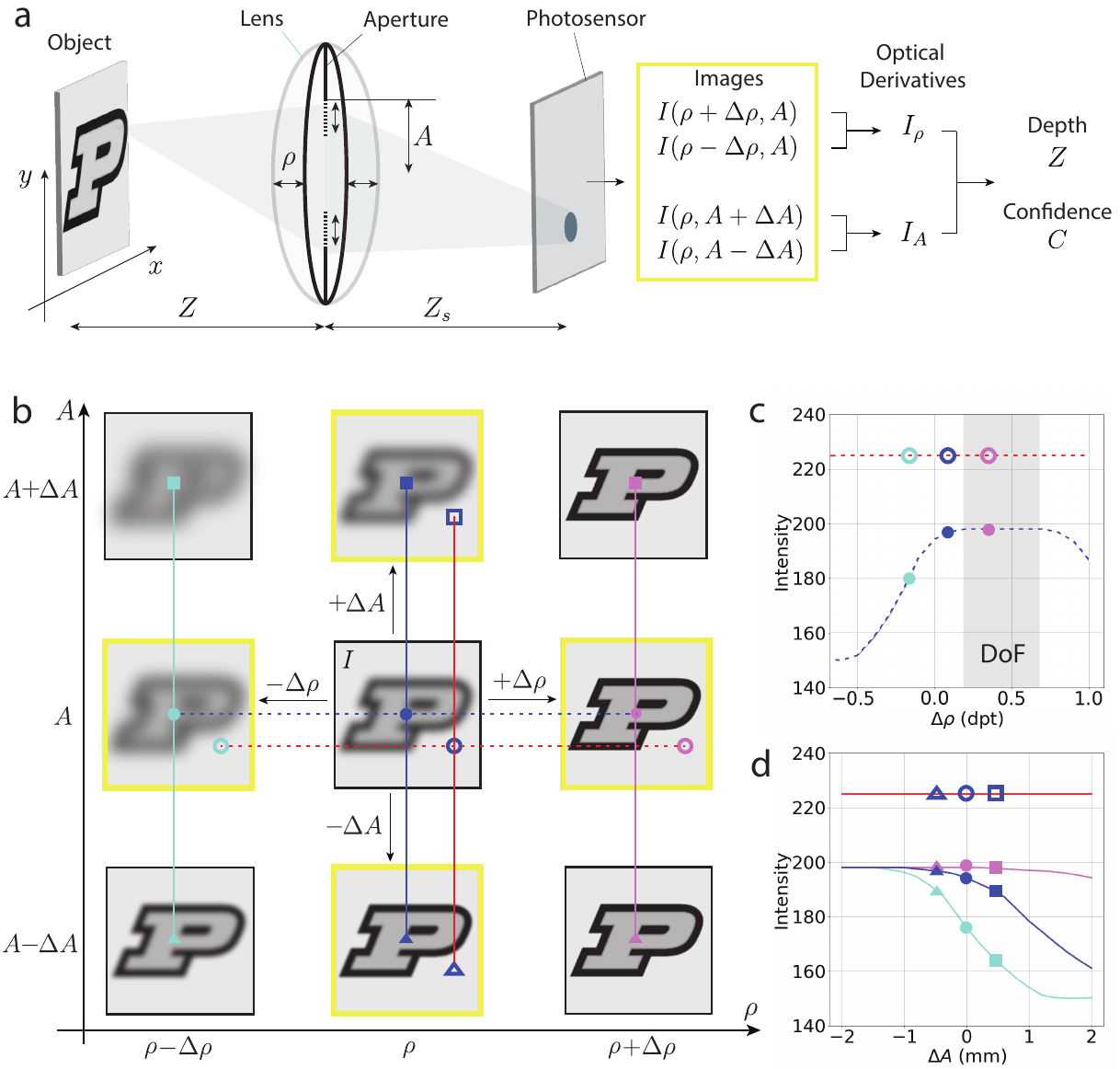}
    \caption{(a) Principle of coupled optical differentiation. Consider a thin lens camera with sensor distance $Z_s$ and adjustable optical power $\rho$ and aperture radius $\mathnormal{A}$. The image it captures is a function of these two optical parameters, $\rho$ and $\mathnormal{A}$, denoted as $I(\rho, \mathnormal{A})$. In this work, we show that the ratio of the optical derivatives, $I_\mathnormal{A} / I_\rho$, reveals the object depth $Z$ at each pixel through closed-form solutions. 
    (b) Images of the same object captured with different optical power $\rho$ and aperture radius $\mathnormal{A}$. By adjusting the optical parameters $\rho, \mathnormal{A}$, the camera can capture images $I$ of the object with different defocus levels. In practice, we can build a system to capture the four highlighted images $I(\rho+\Delta\rho, \mathnormal{A}), I(\rho-\Delta\rho, \mathnormal{A}), I(\rho, \mathnormal{A}+\Delta\mathnormal{A}), I(\rho, \mathnormal{A}-\Delta\mathnormal{A})$ to estimate the optical derivatives $I_\rho$ and $I_\mathnormal{A}$ via finite difference. (c) Pixel intensity vs. optical power  $\rho$. The colored markers indicate the intensities of corresponding image pixels in (b). The intensity varies in textured regions,  e.g., pixel $\bullet$, when the object is out of the depth-of-field (DoF). Meanwhile, the intensity is close to constant in textureless regions, such as at pixel $\circ$. (d) Pixel intensity vs. aperture radius $\mathnormal{A}$. The plot visualizes the intensity of pixel $\bullet$ as a function of aperture radius $\mathnormal{A}$ under three different aperture radii, $\mathnormal{A}-\Delta\mathnormal{A}, \mathnormal{A}, \mathnormal{A}+\Delta\mathnormal{A}$. As the images with optical power $\rho+\Delta\rho$ are in focus (see b), the pixel intensity stays approximately constant w.r.t. the aperture radius $\mathnormal{A}$ (pink curve). }
    \label{fig:principle}
\end{figure*}

\subsection{Image Formation Model}
As shown in Fig.~\ref{fig:principle}a, we consider a thin-lens camera imaging a front-parallel object with spatially-varying intensity $T(x,y)$ located at a constant depth $Z$ from the camera. A pinhole camera with an aperture-to-sensor distance $Z_s$ would capture the all-in-focus perspective image $P(x,y)$:
\begin{align}
    P(x,y;Z) = T\left(-\frac{Z}{Z_s}x,-\frac{Z}{Z_s}y\right).
\label{eq:P}
\end{align}
The image captured by the thin-lens cameras can be modeled with a pinhole projection followed by defocus blur as:
\begin{align}
    I(x,y;Z) = k(x,y;Z) \circledast P(x,y;Z),
\label{eq:ifm}
\end{align}
where $k(x,y;Z)$ is the camera's point spread function (PSF) corresponding to object distance $Z$ and $\circledast$ indicates convolution in $x$ and $y$. We model the PSF as a scaled version of the aperture transmittance profile $\kappa(x,y)$:
\begin{align}
     k(x,y;Z) = \frac{1}{\sigma^2(Z)} ~ \kappa\left(\frac{x}{\sigma(Z)},\frac{y}{\sigma(Z)}\right),
     \label{eq:k}
\end{align}
with the scale $\sigma$ determined by the optical parameters of the camera (aperture radius $\mathnormal{A}$ and optical power $\rho$), as well as the distances $Z_s$ between the aperture and the sensor and $Z$ between the aperture and the scene:
\begin{align}
    \sigma(Z; A, \rho, Z_s) = A + \left(\rho - \frac{1}{Z}\right)~A  ~Z_s.\label{eq:sig}
\end{align}
For objects with slowly varying depth and sparse depth discontinuities, Eq.~\ref{eq:ifm} remains applicable under the patchwise approximation of a front-parallel scene~\citep{guo2019compact, guo2017focal, alexander2018focal}. 

\subsection{Depth Estimation}
\label{secsec:depth}

Our objective is to estimate object depth $Z$ from changes in image brightness $I(x,y)$ caused by changes in defocus level $\sigma$. From Eqs.~\ref{eq:ifm}-\ref{eq:sig} we observe what occurs to the image when optical parameters change, with subscripts indicating partial derivatives:
\begin{align}
\begin{split}
    I_{\mathnormal{A}}(x,y) =& \left[k_{\sigma}(x,y)~ \sigma_\mathnormal{A}\right] \circledast P(x,y) \\
    =& \left(1+\left(\rho - \frac{1}{Z}\right)Z_s\right)\left[k_{\sigma}(x,y) \circledast P(x,y)\right], 
\end{split}\\
\begin{split}
    I_{\rho}(x,y) =& \left[k_{\sigma}(x,y)~ \sigma_{\rho}\right] \circledast P(x,y) \\
    =& A~Z_s~ \left[k_{\sigma}(x,y) \circledast P(x,y)\right].
\end{split}
\end{align}
Prior work has established that the $k_\sigma \circledast P$ term, interpreted as a defocus residual on brightness constancy, can only be observed from spatial derivatives of defocused images if the blur is Gaussian~\citep{alexander2018focal}. However, we note that by comparing the image changes directly across this coupled pair of optical changes, the depth map is revealed immediately for any $k$:
\begin{align}
    \frac{I_A(x,y)}{I_\rho(x,y)} =& \frac{1+\left(\rho - \frac{1}{Z(x,y)}\right)Z_s}{A ~ Z_s} \\
    \rightarrow Z(x,y) =& \frac{Z_s}{Z_s\rho - 1 - \mathnormal{A} Z_s \cdot I_\mathnormal{A}(x,y)/I_\rho(x,y)} \label{eq:rhoSigma}.
\end{align}
  Further,  we no longer require spatial neighborhoods of pixels for depth estimation from spatial derivatives, shrinking both computational cost and sensitivity to local depth variations. 
Hence, we can recover depth with only two divides and a few adds and multiplies per pixel directly from \textit{coupled optical differentiation}.

In practice, these derivatives require physical adjustments to the camera and are measured with finite differences, as in
\begin{align}
    I_{\rho} &= \frac{I(\rho+\Delta \rho) - I(\rho - \Delta \rho)}{2\Delta \rho},
    \label{eq:fin}
\end{align}
for a focus-tunable lens set to $\rho\pm\Delta \rho$. For simplicity, we drop the pixel location $(x,y)$ from now on if the operations are per pixel.
The differentiation of the aperture radius $\Delta A$ is less straightforward to be realized. If the imaging system uses a circular aperture, i.e., the aperture transmittance profile $\kappa(x,y)$ is a 2D pillbox function with radius $\frac{1}{\pi}$:
\begin{align}
    \kappa(x,y) = \frac{1}{\pi}\left(\sqrt{x^2 + y^2} < 1\right),
    \label{eq:pillbox}
\end{align}
we can use a motorized iris to `equivalently' change the aperture radius $A$ by normalizing the brightness of the captured image:
\begin{align}
    I(\mathnormal{A} + \Delta \mathnormal{A}) = \left(\frac{\mathnormal{A}}{\mathnormal{A} + \Delta \mathnormal{A}}\right)^2 \Tilde{I}(\mathnormal{A}+\Delta \mathnormal{A}),
    \label{eq:bright_adjust}
\end{align}
where $\Tilde{I}(\mathnormal{A}+\Delta \mathnormal{A})$ the raw captured image with aperture radius $\mathnormal{A}+\Delta \mathnormal{A}$. In Sec.~\ref{sec:exp}, we demonstrate a sensor prototype that can perform coupled optical differentiation $(\Delta\mathnormal{A}, \Delta \rho)$ using a custom optical assembly with an off-the-shelf motorized iris and deformable lens.  

\subsection{Failure Cases and Mitigation}
\label{secsec:fail}

There are two failure cases for our depth sensing equation
. First, the proposed method fails in image regions that lack spatial variation in intensity, as any triangulation-based method does. This is illustrated in Fig.~\ref{fig:principle}c-d, where the image intensity stays constant with respect to the change of optical parameters $\rho$ and $\mathnormal{A}$ at a pixel in the textureless region. Thus, both image derivatives $I_\rho$ and $I_\mathnormal{A}$ in Eq.~\ref{eq:rhoSigma} becomes zero. Fortunately, this situation can be detected directly from the values of image derivatives $I_\rho$ and $I_\mathnormal{A}$, which can inform confidence of the depth estimation to filter out bad pixels. We seek a per-pixel, computationally inexpensive confidence metric high in regions of strong image derivative signals and vice versa.

Second, the finite difference estimation of image derivatives becomes inaccurate when the captured images are close to focus. There are several causes of this phenomenon, which can be witnessed in Fig.~\ref{fig:principle}c-d. When changing the optical power $\rho$, the pixel intensity $I$ is constant when the object is in the depth-of-field (Fig.~\ref{fig:principle}c.) Besides, the pixel intensity $I$ also remains constant w.r.t. the aperture radius $\mathnormal{A}$ if the object is in focus (Fig.~\ref{fig:principle}d.) This indicates that both image derivatives $I_\rho$ and $I_\mathnormal{A}$ go to zero when the object is in focus, and Eq.~\ref{eq:rhoSigma} degenerates. 

We can identify pixels with failed depth estimation using a simple, per-pixel confidence metric based on the derivative magnitude:
\begin{align}
    C_{(\text{Eq.}~\ref{eq:rhoSigma})} &=  I_{\rho}^2. \label{eq:rhoSigma_C} 
\end{align}
and filter these pixels out using a pre-determined, fixed confidence threshold $C_{\text{thre}}$: 
\begin{align}
    Z = \begin{cases}
        Z, \quad C > C_{\text{thre}} \\
        \text{unconfident}, \quad \text{otherwise}.
    \end{cases}
\end{align}
We show in Sec.~\ref{secsec:sim-conf} and Sec.~\ref{secsec:quant} the effectiveness of this simple confidence metric to filter out erroneous depth estimations in both simulation and real experiments.

\begin{figure*}
    \centering
    \includegraphics[width=\linewidth]{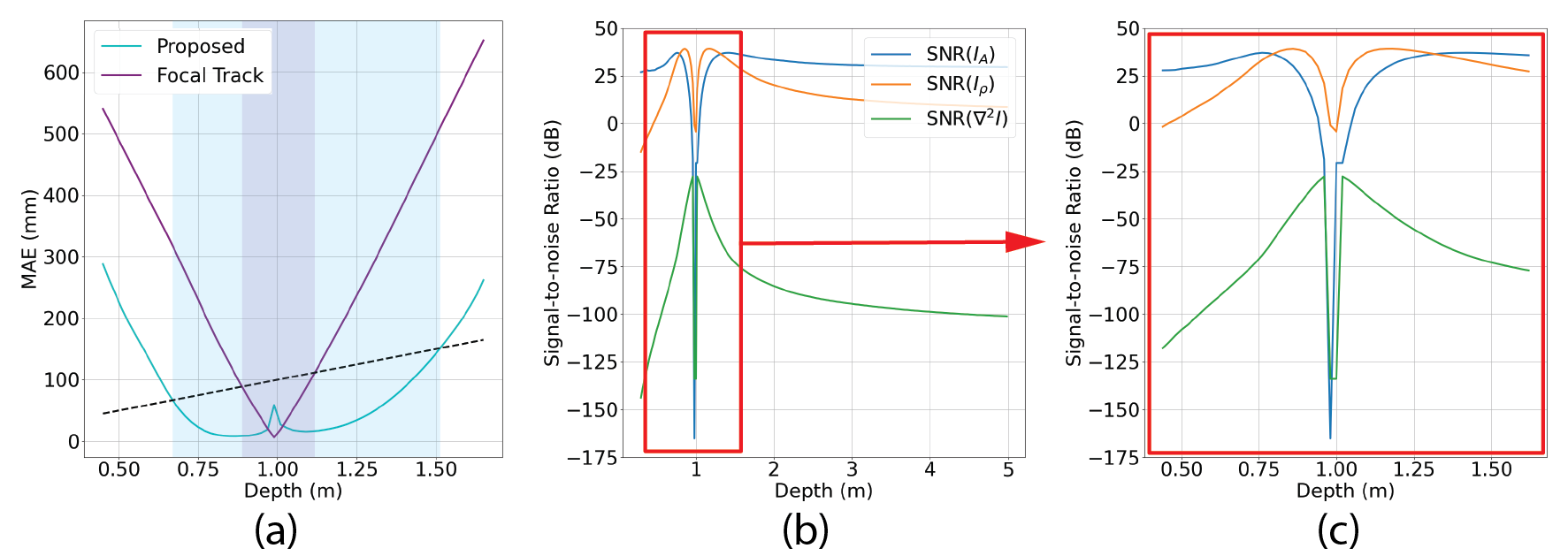}
    \caption{Working range of the proposed method and Focal Track~\citep{guo2017focal}. (a) Mean absolute error (MAE) as a function of depth, with the black dashed line marking the 10\% of the depth value. The highlighted regions indicate the working ranges of both methods. Throughout the paper, we define the working range as where the MAE is smaller than $10\%$ of the true depths. The proposed method's working range is four times that of Focal Track. (b) Signal-to-noise ratio (SNR) of optical derivatives $I_\mathnormal{A}, I_\rho$, and the spatial derivative $\nabla^2 I$. The optical derivatives generally have a significantly larger SNR than the spatial derivative $\nabla^2 I$, which explains the higher accuracy and longer working range of the proposed method, where only the optical derivatives $I_\mathnormal{A}, I_\rho$ have been used. Meanwhile, Focal Track leverages the spatial derivative $\nabla^2 I$ for depth estimation. (c) The enlarged portion of (b). The SNRs of optical derivatives $I_\mathnormal{A}, I_\rho$ drop when the object is in focus, i.e., at around 1 m, as explained in Sec.~\ref{secsec:fail}. This accounts for the proposed method's sudden MAE increase at 1 m in (a). }
    \label{fig:working_range}
\end{figure*}

\section{Analysis}

This section comprehensively analyzes the depth from coupled optical differentiation theory using computer-synthesized data, including the working range, confidence, and optimal aperture transmittance profile. 
We simulate an ideal thin-lens camera, as described in Fig.~\ref{fig:principle}, imaging front-parallel objects with textures sampled from a natural texture dataset~\citep{dana1999reflectance} throughout all studies presented in this section. Without loss of generality, we adopt a specific set of optical parameters in simulation that approximately match the real prototype to be presented in Sec.~\ref{sec:exp}.

\subsection{Working Range Advantage}
\label{secsec:snr}

One major advantage of depth from coupled optical differentiation is the larger working range compared to previous DfD algorithms that leverage spatial derivatives of images. For example, Focal Track~\citep{guo2017focal} uses a similar depth sensing equation as Eq.~\ref{eq:rhoSigma}:
\begin{align}
    Z = \frac{\mathnormal{A}^2 Z_s^2}{\mathnormal{A}^2Z_s(Z_s \rho + 1) - I_\rho / \nabla^2 I},
    \label{eq:focal_track}
\end{align}
but it requires the second-order spatial derivative of the image, $\nabla^2 I$. Fig.~\ref{fig:working_range}a shows the depth prediction accuracy of using our method (Eq.~\ref{eq:rhoSigma}) and Focal Track (Eq.~\ref{eq:focal_track}) with the same optical configurations and noise level. Ours achieves a much smaller mean absolute error (MAE) for almost all depths. 

Throughout this paper, we evaluate the depth sensing performance using the \textit{working range}, defined as the set of scene depths over which the MAE of recovered depth is less than 10\% of the ground truth values. We highlight the working range of both methods in Fig.~\ref{fig:working_range}a: our method
achieves a four times larger working range than Focal Track.
(80 cm vs 20 cm.)


\begin{figure*}[t!]
    \centering
\includegraphics[width=1.0\textwidth]{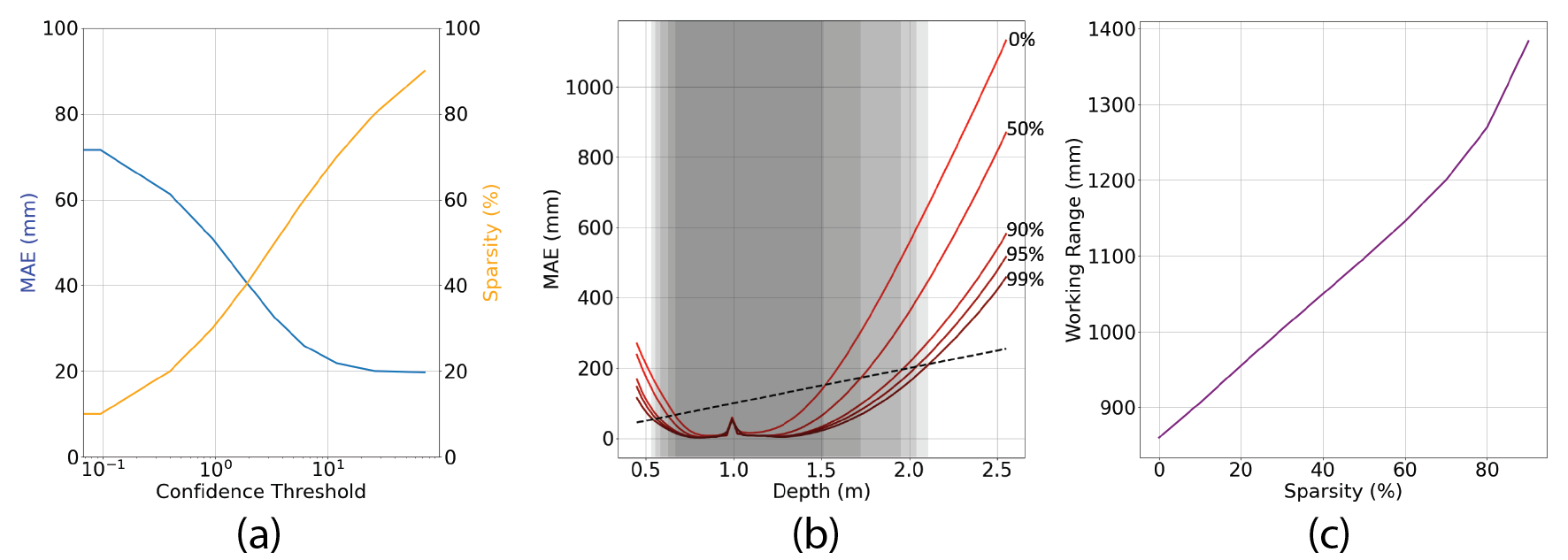}
    \caption{Effect of confidence. (a) The MAE of predicted depth (blue) and the sparsity (yellow) as a function of the confidence threshold. We filter out depth predictions by comparing their corresponding confidence values with a predefined confidence threshold. As the confidence threshold increases, only pixels with higher confidence values remain, and the \textit{sparsity} of the depth map increases. The blue curve clearly shows the decrease of the MAE when increasing the confidence threshold, which suggests the effectiveness of the confidence metric. (b) MAE as a function of true depth with different confidence thresholds. By increasing the confidence threshold, the sparsity increases and the MAE generally drops for all depths. We label the overall sparsity and highlight the working range for each curve. (c) Working range as a function of overall sparsity, a proxy of confidence threshold. }
    \label{fig:ausc}
\end{figure*}

Our method's higher accuracy and more extended working range can be explained using the signal-to-noise ratios (SNR) of the estimated image derivatives $I_\rho, I_\mathnormal{A}, $ and $\nabla^2 I$ in Eq.~\ref{eq:rhoSigma} and Eq.~\ref{eq:focal_track} in the presence of noise. Assuming sufficient photons when capturing the images, we use the following image noise model~\citep{hasinoff2021photon}:
\begin{align}
    I = I^* + \sqrt{I^*}\epsilon.
    \label{eq:noise}
\end{align}
The symbol $I^*$ denotes the noiseless image, and the random variable $\epsilon$ follows the standard normal distribution $\epsilon \sim \mathcal{N}(0,\frac{1}{\lambda})$, where $\lambda$ is the photon per brightness level of the camera system. Then, we calculate the SNR of the estimated image derivatives at every pixel via the following equations:
\begin{align}
    \text{SNR}(I_\rho) &= \left|\frac{I^*_\rho}{\frac{I(\rho+\Delta \rho) - I(\rho-\Delta\rho)}{2\Delta\rho} - I^*_\rho}\right|,  \\
    \text{SNR}(I_\mathnormal{A}) &= \left|\frac{I^*_\mathnormal{A}}{\frac{I(\mathnormal{A}+\Delta \mathnormal{A}) - I(\mathnormal{A}-\Delta \mathnormal{A})}{2\Delta \mathnormal{A}} - I^*_\mathnormal{A}}\right|, \\
    \text{SNR}(\nabla^2 I) &= \left|\frac{L\circledast I^*}{L \circledast I - L \circledast I^*}\right|
\end{align}
where the terms $I_\rho^*$ and $I^*_\mathnormal{A}$ are true image derivatives without using finite differences, $L$ is the finite Laplacian filter, and $\circledast$ represents 2D convolution.
Fig.~\ref{fig:working_range}b plots the average SNR of the estimated image derivatives $I_\rho$, $I_\mathnormal{A}$, and $\nabla^2 I$ at different depths. Optical derivatives $I_\rho$, $I_\mathnormal{A}$ have a much higher SNR than the spatial derivative $\nabla^2 I$ over an extended depth range. This illustrates the advantage of the proposed method, which only leverages optical derivatives, compared to previous DfD algorithms that all use spatial derivatives~\citep{subbarao1994depth, alexander2018focal, guo2017focal, guo2019compact}. 

\subsection{Effect of Confidence}
\label{secsec:sim-conf}


The simple confidence metric we proposed in Eq.~\ref{eq:rhoSigma_C} effectively masks out failed depth predictions. Fig.~\ref{fig:ausc}a visualizes the sparsification plot using the confidence metric. The figure shows the mean absolute error (MAE) of depth predictions at all object distances when a portion of the least confident pixels below a threshold is discarded. We define the portion of the discarded pixel as the \textit{sparsity}. The higher the threshold is, the higher the sparsity is, and the higher the confidence values of the remaining pixels are. As demonstrated in Fig.~\ref{fig:ausc}a, the overall MAE gradually reduces from around 70 mm to 20 mm as the sparsity increases, proving that depth predictions with higher confidence values generally have higher accuracy. Fig.~\ref{fig:ausc}b shows the MAE as a function of each true depth at different sparsities. The depth prediction error is universally lower at each true depth when increasing the sparsity, in other words, the confidence threshold. Furthermore, we plot the working range as a function of sparsity in Fig.~\ref{fig:ausc}c and witness a monotonic increase in the working range as the sparsity grows. All these results clearly show the effectiveness of the confidence metric in predicting the reliability of the depth estimation at each pixel.


\begin{figure*}
    \centering
    \includegraphics[width=1.0\textwidth]{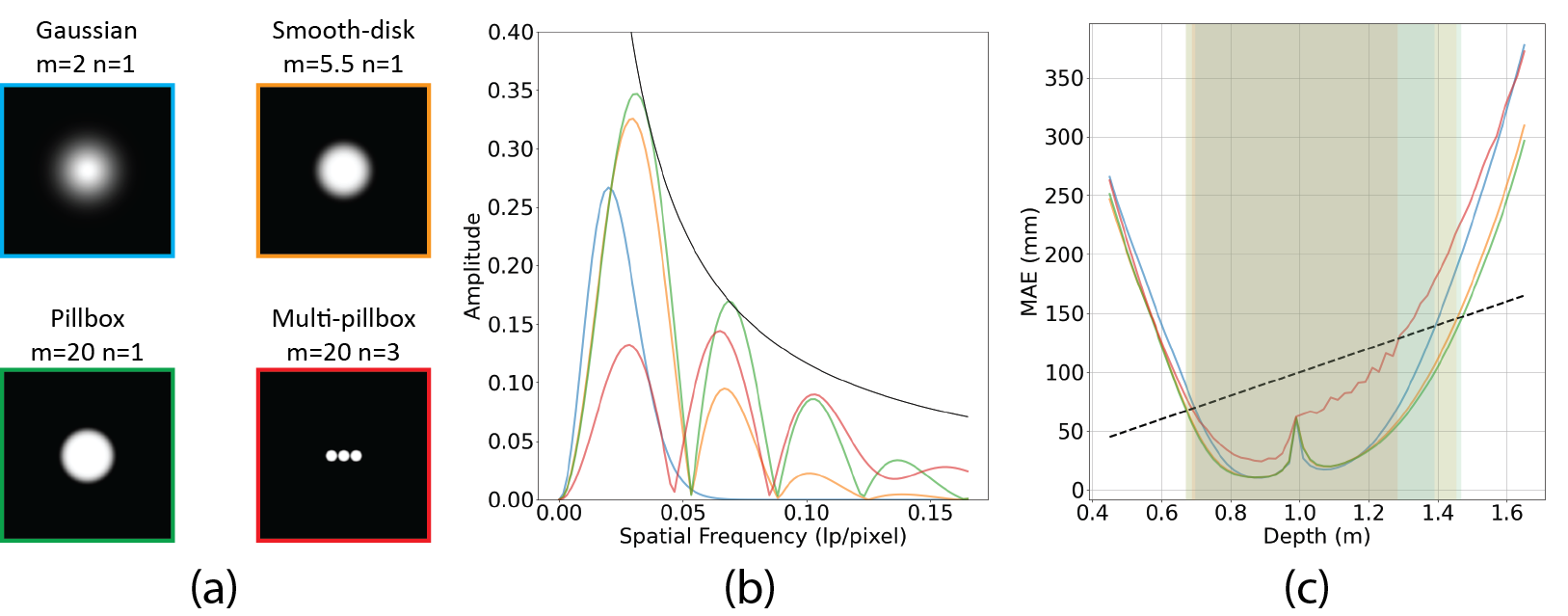}
    \caption{Aperture transmittance profile analysis. (a) Four different apertures parameterized using Eq.~\ref{eq:aper_code_eq}. The colors of the boxes indicate the corresponding curves in (b) and (c). (b) Amplitude spectrum of the finite optical derivative of the PSFs, $k(\rho+\Delta\rho) - k(\rho-\Delta\rho)$, for each aperture transmittance profile at a specific depth. The black curve indicates the 1/f statistics of natural textures. The pillbox aperture (green) achieves the highest overall amplitude, with the smooth disk being the second. This amplitude spectra relationship is typical at other depths within the working range. (c) The MAE of different aperture transmittance profiles. Consistent with the conclusion of (b), the pillbox aperture achieves the lowest MAE at a wide range of depths.}
    \label{fig:aperture-code}
\end{figure*}

\subsection{Aperture Transmittance Profile}
\label{secsec:psf}



One significant advantage of the coupled optical differentiation theory compared to previous DfD theories, such as depth from differential defocus~\citep{alexander2019theory}, is that it does not require a specific aperture transmittance profile. In theory, the aperture transmittance profiles do not affect depth estimation accuracy because they are canceled out during the calculation process, as seen in Eq.~\ref{eq:rhoSigma}. However, different aperture transmittance profiles will result in different depth estimation accuracy in practice, as the derivatives are approximated by finite difference, and the SNR of the approximation depends on the shape of the PSFs. This section explores how different aperture transmittance profiles affect depth estimation accuracy.

\begin{figure*}
    \centering
    \includegraphics[width=0.9\linewidth]{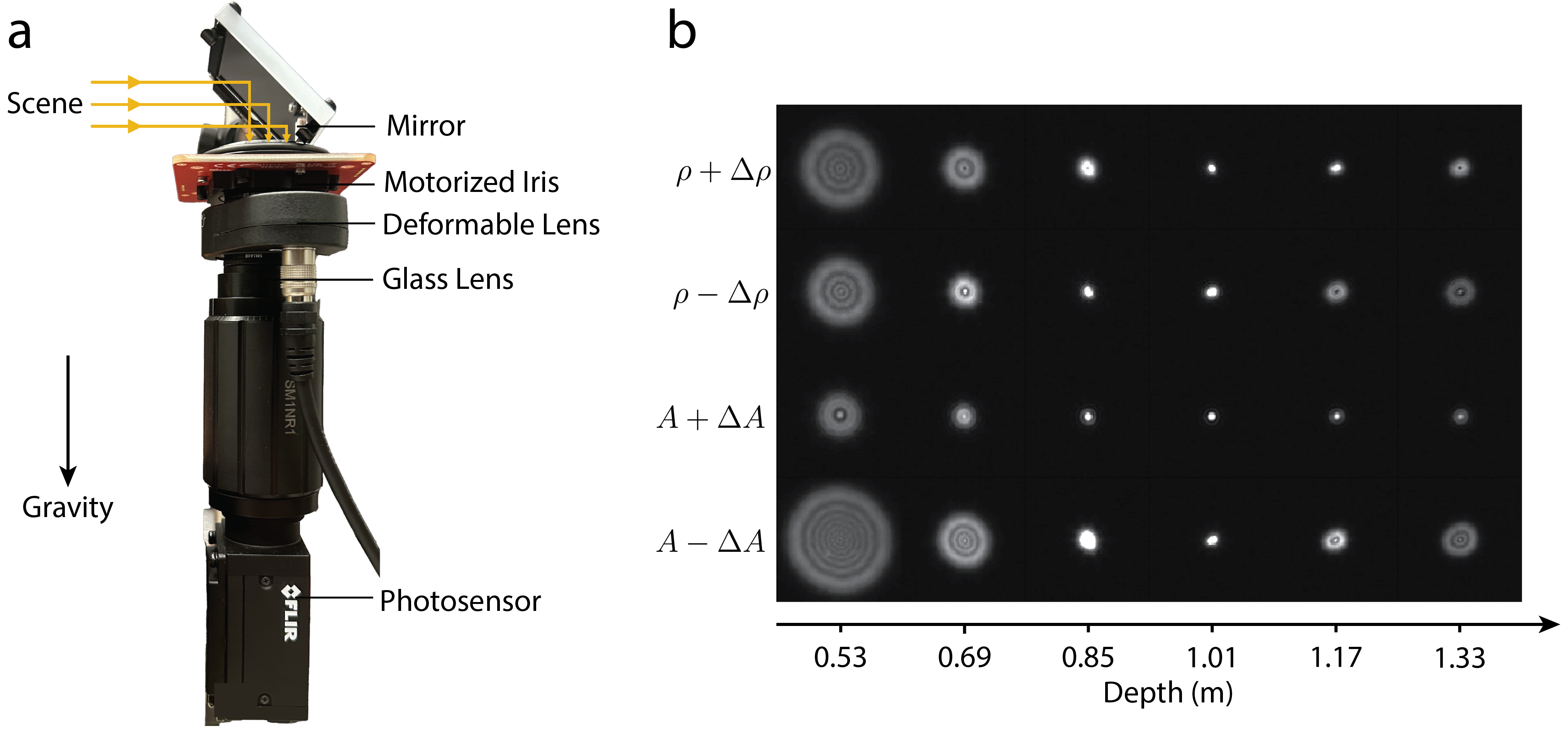}
    \caption{Prototype system. (a) Picture of the optical setup. The optics consist of a Thorlabs ELL15K motorized iris, which can dynamically adjust its diaphragm between 1mm and 12mm, and an Optotune EL-16-40-TC-VIS-5D-C electric tunable lens that can adjust $\rho$ between -2 dpt to 3 dpt. We place a glass lens to adjust the system's overall working range. The photosensor is the FLIR Grasshopper GS3-U3-23S6M-C, configured to capture 16-bit, $480\times300$ images. (b) PSFs of the four captured images $I(\rho+\Delta\rho, \mathnormal{A}), I(\rho-\Delta\rho, \mathnormal{A}), I(\rho, \mathnormal{A}+\Delta \mathnormal{A}), I(\rho, \mathnormal{A}-\Delta \mathnormal{A})$ at different depths.}
    \label{fig:system_photo}
\end{figure*}

We define a general formula that models the family of aperture transmittance profiles we study in this section:
\begin{align}
\begin{split}
        \kappa&(x,y; m,n) = \\ &\sum_{i=1}^n\exp{\left[-\left(\frac{(x-x_i)^2+(y-y_i)^2}{2\sigma_i^2}\right)^{m/2}\right]},
    \label{eq:aper_code_eq}
\end{split}
\end{align}
where $n$ defines the number of blobs in the transmittance profile and $m$ is the smoothness of the blobs. Sample profiles that can be modeled using this formula are shown in Fig.~\ref{fig:aperture-code}a, including Gaussian, pillbox, smooth-disk, and multi-pillboxes. For each aperture transmittance profile, we analyze the amplitude spectrum of the corresponding PSF's finite optical derivative, $k_\rho \approx k(\rho+\Delta\rho) - k(\rho-\Delta\rho)$. As the optical derivative $I_\rho$ is mathematically the convolution of $k_\rho$ with the pinhole image, the amplitude spectrum of different $k_\rho$ indicates the power spectrum of the estimated $I_\rho$. As shown in Fig.~\ref{fig:aperture-code}b, the pillbox aperture achieves the highest overall amplitude spectrum in $k_\rho$ and, interestingly, is mainly aligned with the 1/f relationship of natural textures (black solid curve). This is consistent with the depth estimation accuracy of different aperture transmittance codes shown in Fig.~\ref{fig:aperture-code}c, where the pillbox aperture achieves the lowest MAE at most depths. This evidence empirically suggests the optimality of the pillbox aperture transmittance profile within the family we studied. It validates the prototype sensor design in Sec.~\ref{sec:exp} that uses a pillbox aperture.

\section{Prototyping \& Experimental Results}
\label{sec:exp}

\subsection{Optical System}


We design and build an imaging system that can perform the coupled optical differentiation described in Eq.~\ref{eq:rhoSigma}. The optical assembly of the system consists of a deformable lens and a motorized iris, which can dynamically adjust the optical power of the system $\rho$ and the aperture dimension $\mathnormal{A}$, respectively, and a fixed focal length lens to offset the overall optical power of the system. See Fig.~\ref{fig:system_photo}a. The photosensor of the system is FLIR Grasshopper GS3-U3-23S6M-C. The original resolution of the sensor is 1920$\times$1200. We configured the photosensor to bin every 4$\times$4 pixels so that it outputs 16-bit, 480$\times$300-pixel monochrome images. This way, the readout can achieve the lowest shot noise and discretization noise in the captured images. As shown in Fig.~\ref{fig:system_photo}a, we assemble the optical system vertically to reduce the optical aberration of the deformable lens caused by gravity and use a mirror to adjust the system's field of view.


\subsection{Calibration}

We identify two primary optical aberrations of the optical system that affect the depth sensing accuracy, including the non-uniform background light in the images and the magnification shifting when adjusting the optical power of the deformable lens. We briefly describe the calibration and attenuation process for these two artifacts. In addition, we also calibrate the image noise according to the noise model in Eq.~\ref{eq:noise}. 



\subsubsection{Brightness Registration} 
We observe smoothly varying background light in images captured using our system. In particular, the background light varies when adjusting the aperture dimension $\mathnormal{A}$ of the motorized iris, while it remains fixed when the optical power $\rho$ changes. Thus, this aberration significantly impacts the estimation of $I_\mathnormal{A}$, but the estimation of $I_\rho$ is unaffected since the aberration can be canceled during the finite difference. We notice the background light is typically smoothly varying, so we propose to attenuate it via the following procedure:
\begin{align}
    I_\mathnormal{A} = \Tilde{I}_\mathnormal{A} - B *  \Tilde{I}_\mathnormal{A},
    \label{eq:background-removal}
\end{align}
where $I_\mathnormal{A}$ denotes the clean derivative and $\Tilde{I}_\mathnormal{A} = I(\rho, \mathnormal{A}+\Delta \mathnormal{A}) - I(\mathnormal{A}-\Delta \mathnormal{A})$ represent the corrupted derivatives. In our experiment, we set the averaging kernel $B$ as a 2D box filter with dimension $21\times 21$, as 2D box filtering is separable and can be implemented efficiently using only five FLOPOP~\citep{nakamura2017fast}. 

\begin{figure}
    \centering
    \includegraphics[width=\linewidth]{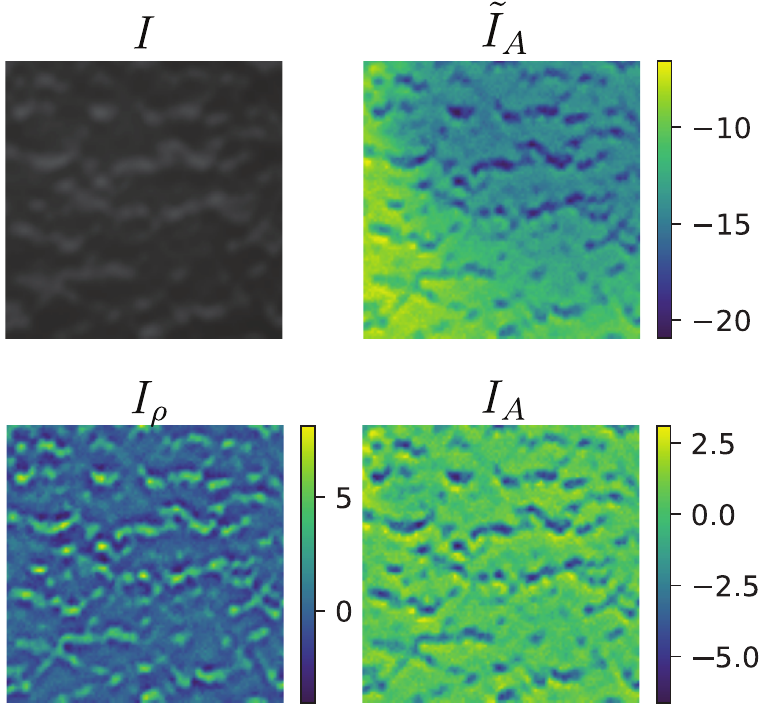}
    \caption{Brightness registration. The image $I$ is a sample captured image of a front-parallel textured object. The corrupted image derivative $\Tilde{I}_\mathnormal{A}$ is directly calculated via finite difference (Eq.~\ref{eq:fin}). The non-uniform background light causes a smoothly changing offset in $I_\mathnormal{A}$, which contaminates the depth estimation. After removing the non-uniform background lighting via Eq.~\ref{eq:background-removal}, the clean image derivative $I_\mathnormal{A}$ visually matches the intensity profile of the optical derivative $I_\rho$. }
    \label{fig:I_A_calib}
\end{figure}







\subsubsection{Geometric Alignment}
\begin{figure}
    \centering
    \includegraphics[width=\linewidth]{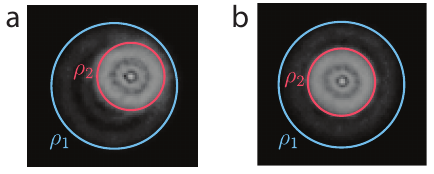}
    \caption{Geometric alignment. (a) The images of a fixed point source captured at two different optical powers, $\rho_1$ and $\rho_2$, are overlaid to show the magnification shifting of the optical system when the optical power $\rho$ varies. We circle the contour of the images to highlight the shift. (b) After the geometric alignment, the two overlaid images appear concentric, which indicates the magnification shifting has been mitigated. }
    \label{fig:I_rho_calib}
\end{figure}

We notice another aberration affecting depth sensing performance: magnification shifting. As illustrated in Fig.~\ref{fig:I_rho_calib}a, the magnification of the image slightly changes as the optical power $\rho$ varies, which causes the image of a fixed point source to move its center position. Interestingly, the magnification shifting can be ignored when the aperture radius varies or the object depth changes. Thus, we only need to geometrically align images $I(\rho+\Delta\rho, \mathnormal{A})$ and $I(\rho-\Delta\rho, \mathnormal{A})$ to the other two images $I(\rho, \mathnormal{A}+\Delta \mathnormal{A}), I(\rho, \mathnormal{A}-\Delta \mathnormal{A})$.

To model the magnification shifting, we define the center of a fixed point source's image on the photosensor $\boldsymbol{x} = [x,y]^T$ as a function of the deformable lens' optical power $\rho$ and the center of the image $\boldsymbol{x}_0$ at a reference optical power $\rho_0$:
\begin{align}
    \boldsymbol{x}(\rho, \boldsymbol{x}_0) = [\boldsymbol{\lambda}, A, I]
    \left[\begin{array}{c}
         \rho  \\
         \rho \boldsymbol{x}_0 \\
         \boldsymbol{x}_0
    \end{array}\right], 
    \label{eq:mag-model}
\end{align}
where the matrix $A\in \mathbb{R}^{2\times 2}$ and the vector $\boldsymbol{\lambda}\in \mathbb{R}^{2\times 1}$ are the parameters to be calibrated, and the matrix $I\in \mathbb{R}^{2\times 2}$ is the identity matrix. By placing a point source at different positions $i=1,2,\cdots$ and capturing images under optical powers $\rho_j, j=1,2,\cdots$, we can measure the centers of the point source's image $\boldsymbol{x}^i_j$ and fit the magnification model (Eq.~\ref{eq:mag-model}) via:
\begin{align}
    \Tilde{A}, \Tilde{\boldsymbol{\lambda}} =\underset{A, \boldsymbol{\lambda}}{\text{arg min}} \sum_{i,j}\lVert \boldsymbol{x}(\rho_j, \boldsymbol{x}^i_0) - \boldsymbol{x}^i_j\rVert^2.
\end{align}
After calibrating the parameters of the magnification shifting $A, \boldsymbol{\lambda}$, we can determine a per-pixel correspondence between images captured with different optical powers, $\rho_1$ and $\rho_2$, via: 
\begin{align}
\begin{split}
        \boldsymbol{x}_2&(\boldsymbol{x}_1) = \\ &(\rho_2 A + I)(\rho_1 A + I)^{-1} (\boldsymbol{x}_1 - \rho_1 \boldsymbol{\lambda}) + \rho_2 \boldsymbol{\lambda},
    \label{eq:pixel-mapping}
\end{split}
\end{align}
where $\boldsymbol{x}_1$ and $\boldsymbol{x}_2$ are corresponding pixels in two images captured with optical powers $\rho_1$ and $\rho_2$. As the magnification shifting is fixed after the system is assembled, we can pre-define a bilinear interpolation model to align the images. For example, given an unaligned image $\Tilde{I}_2$ and a target image $I_1$. The operation is:
\begin{align}
    I_2(\boldsymbol{x}) = \sum_{k=1}^4 w_{2,k}(\boldsymbol{x}) \Tilde{I}_2(\tilde{\boldsymbol{x}}_{2,k}(\boldsymbol{x})),
    \label{eq:interpolate}
\end{align} 
where $I_2$ is the aligned image of $\Tilde{I}_2$. The pixels $\tilde{\boldsymbol{x}}_{2,k}(\boldsymbol{x}), k=1,\cdots,4$ are the four neighboring pixels of position $\boldsymbol{x}_2(\boldsymbol{x})$, which corresponds to pixel $I_1(\boldsymbol{x})$ in the unaligned $\Tilde{I}_2$ following Eq.~\ref{eq:pixel-mapping}. The coefficients $w_{2,k}, k=1,\cdots,4$ are the bilinear weights. We can precalculate the store the corresponding pixel locations $\tilde{\boldsymbol{x}}_{2,k}(\boldsymbol{x})$ and weights $w_{2,k}$ for each $\boldsymbol{x}$. During inference, the geometric alignment (Eq.~\ref{eq:interpolate}) becomes a linear combination for every pixel of $I_2(\boldsymbol{x})$. Fig.~\ref{fig:I_rho_calib}b overlays images of a fixed point source captured at different optical powers after the geometric alignment. Compared to before the alignment (Fig.~\ref{fig:I_rho_calib}a), the aligned images appear concentric, demonstrating the effectiveness of the geometric alignment. 

\subsubsection{Noise Level}

We calibrate the photon per brightness level, $\lambda$, of the noise model listed in Eq.~\ref{eq:noise}. We capture 100 images of a static scene, $I_i(x,y), i=1,2,\cdots, 100$ with fixed exposure time and gain of the photosensor,. Assuming the true brightness can be accurately approximated using the empirical mean of the 100 images, we can calculate the maximum likelihood estimation of the photon per brightness level $\lambda$ via:
\begin{align}
    \begin{split}
        \arg\min_\lambda \sum_{x,y} \sum_i \log(\Bar{I}(x,y)) + \log(\lambda) + \\
        \frac{1}{\lambda \Bar{I}(x,y)}\left(I_i(x,y) - \Bar{I}(x,y)\right)^2,
    \end{split}
\end{align}
where $\Bar{I}(x,y)$ is the empirical mean of the 100 images, $\Bar{I}(x,y) = \sum_i I_i(x,y)/100$. We use the calibrated photon per brightness level $\lambda$ in subsequent simulations to determine the optical parameter selections. For our system, the calibrated photon per brightness level $\lambda$ is 0.9375 for the 16-bit, 480$\times$300 images.

\subsection{Parameter Selection}

\subsubsection{Optimal Finite Difference Steps}
\label{secsecsec:opt-diff-step}

The proposed system measures the optical derivatives $I_\rho$ and $I_\mathnormal{A}$ via finite difference (Eq.~\ref{eq:fin}) from four captured images: $I_\rho = I(\rho+\Delta\rho, \mathnormal{A}) - I(\rho-\Delta\rho, \mathnormal{A})$, $I_\mathnormal{A} = I(\rho, \mathnormal{A}+\Delta \mathnormal{A}) - I(\rho, \mathnormal{A}-\Delta \mathnormal{A})$. The finite difference steps $\Delta \rho$, $\Delta \mathnormal{A}$ are hyperparameters that need to be determined in advance. The larger $\Delta \rho$ and $\Delta \mathnormal{A}$, the higher the intensity of $I_\rho$ and $I_\mathnormal{A}$ will be, which will be less susceptible to noise. Meanwhile, a large $\Delta \rho$ and $\Delta \mathnormal{A}$ will cause the finite difference to deviate from the ground truth derivatives. The optimal $\Delta \rho$ and $\Delta \mathnormal{A}$ balance this tradeoff.


We optimize the finite difference steps $\Delta \rho$ and $\Delta \mathnormal{A}$ using synthetic images with ground truth depth maps. The images are simulated with optical parameters and the noise level of the prototype system. We calculate the depth maps using Eq.~\ref{eq:rhoSigma} with optical derivatives $I_\rho$ and $I_\mathnormal{A}$ estimated from finite difference. The objective function minimizes the depth prediction error:
\begin{align}
\footnotesize
 \Delta \Tilde{\rho}, \Delta \Tilde{\mathnormal{A}} &= \underset{\Delta\rho, \Delta \mathnormal{A}}{\arg\min}  \sum_{x,y,l}|Z^{l,*}(x,y) - Z^l(x,y; \Delta\rho, \Delta \mathnormal{A})|, \\
 &\text{s.t. } 0<\Delta \rho < \Delta\rho_m,  0 < \Delta \mathnormal{A} < \Delta\mathnormal{A}_m,
\end{align}
where $Z^{l,*}$ and $Z^l$ indicates the $l$th true and predicted depth map, and $\Delta \rho_m$ and $\Delta \mathnormal{A}_m$ represent the maximum feasible finite difference of the prototype, $\Delta\rho_m = 3\;\text{dpt}$ and $\Delta \mathnormal{A}_m = 1\;\text{mm}$. The optimization converges to $\Delta \Tilde{\rho} = 0.06\;\text{dpt}$ and $\Delta \Tilde{\mathnormal{A}} = 1\;\text{mm}$, and we adopt these parameters to capture all remaining results in this manuscript. We measure and visualize the PSFs of the four images $I(\rho+\Delta\Tilde{\rho}, \mathnormal{A}), I(\rho-\Delta\Tilde{\rho}, \mathnormal{A}), I(\rho, \mathnormal{A}+\Delta \Tilde{\mathnormal{A}}), I(\rho, \mathnormal{A}-\Delta \Tilde{\mathnormal{A}})$ at different depths in Fig.~\ref{fig:system_photo}b.

\subsubsection{Derivative aggregation}
\label{eq:comp-acc}

\begin{figure}
    \centering
    \includegraphics[width=0.7\linewidth]{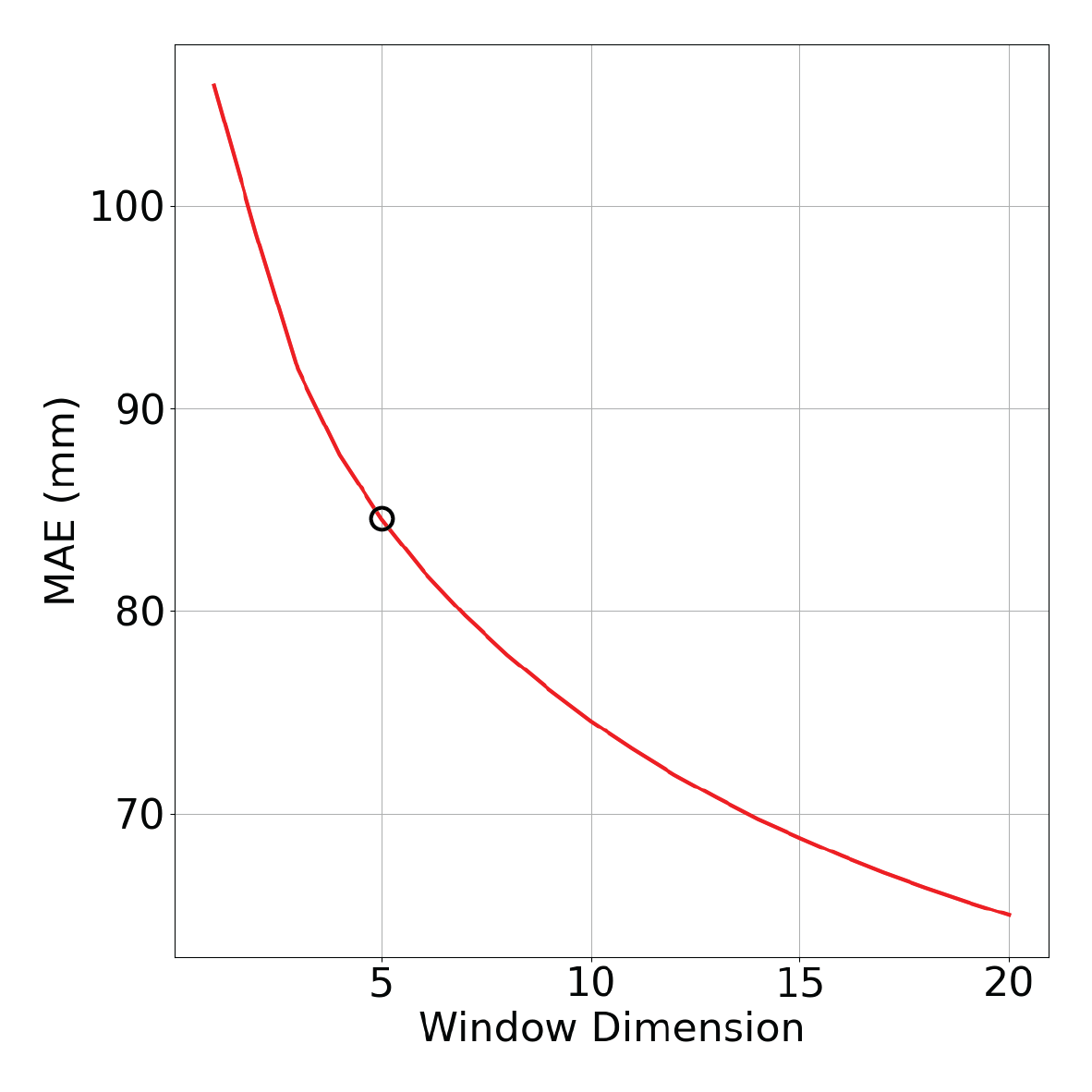}
    \caption{MAE of depth estimation as a function of window dimension in Eq.~\ref{eq:rhoSigma_w}. The elbow point of the curve is marked by the black hollow circle, which balances the depth accuracy and the computational complexity, as both increase with the window dimension.}
    \label{fig:mn-window}
\end{figure}

Eq.~\ref{eq:rhoSigma} estimates the object depth $Z$ corresponding to each pixel $(x,y)$ only using the image derivatives $I_\rho, I_\mathnormal{A}$ at that pixel. In the presence of significant image noise, the depth value at a pixel $(x_0, y_0)$ can be solved more accurately via least square fitting by aggregating image derivatives of pixels within a small window $W$ centered at $(x_0, y_0)$, assuming the depth value remains constant within $W$: 
\begin{align}
\footnotesize
\begin{split}
    &Z(x_0, y_0) = \\ &\frac{\sum_{x,y\in W} I_\rho(x,y) \left((Z_s \rho + 1) I_\rho(x,y) - \mathnormal{A} Z_s I_\mathnormal{A}(x,y)\right)}{\sum_{x,y \in W} \left((Z_s \rho + 1) I_\rho(x,y) - \mathnormal{A} Z_s I_\mathnormal{A}(x,y)\right)^2}.
    \label{eq:rhoSigma_w}
\end{split}
\end{align}
Fig.~\ref{fig:mn-window}b visualizes the trade-off between the depth estimation error and the window dimension in Eq.~\ref{eq:rhoSigma_w}: An increased window dimension improves the depth accuracy, but also increases the computational cost and reduces the spatial resolution of the depth map. We detect the elbow point of the curve and use the corresponding window dimension, $5\times 5$, in the prototype, which balances the accuracy and computational cost. 

\subsection{Computation}

The pseudocode in Algorithm~1 shows our implementation of the depth from coupled optical differentiation, which takes 14 to 36 FLOPOPs, depending on whether to execute certain optional operations. The FLOPOP number considers all output pixels, including those discarded by the confidence metric.
Instead of using the mathematical equation in Eq.~\ref{eq:rhoSigma_w} to calculate depth from the optical derivatives $I_\rho$ and $I_\mathnormal{A}$, we use a pre-determined look-up table (Line 8) to map the ratio of optical differentiation,  $I_{\text{num}}/I_{\text{den}}$, to the predicted depth $Z$ at each pixel. This is due to the challenge to accurately determine the optical parameters, e.g., aperture-to-sensor distance $Z_s$ and aperture scale $\mathnormal{A}$. We build the look-up table by learning the relationship between $I_{\text{num}}/I_{\text{den}}$ and the depth from real data. The data consists of images of a front-parallel texture at a series of known depths, which is effectively a supervised dataset between $I_{\text{num}}/I_{\text{den}}$ and the corresponding depth $Z$. Then, we digitize the ratio of all pixels, $I_{\text{num}}/I_{\text{den}}$, into discrete bins and calculate the median depth of all pixels within each bin. The look-up table maps each digitized ratio to the corresponding depth values. As shown in Line 8 of Algorithm~1, we first digitize the ratio $I_{\text{num}}/I_{\text{den}}$ of each pixel and then use the look-up table to determine the depth value during inference.

\begin{figure}[h!]
    \centering
    \vspace{-0.1in}
    \includegraphics[width=\linewidth]{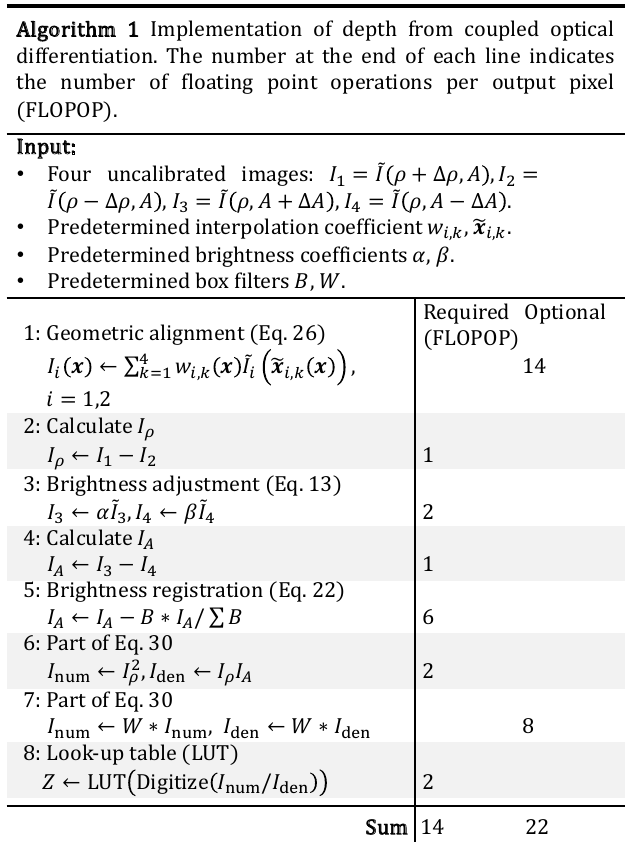}
    \vspace{-0.2in}
\end{figure}

\subsection{Results}
\label{secsec:quant}

\begin{figure*}[t!]
    \centering
    \includegraphics[width=1.0\linewidth]{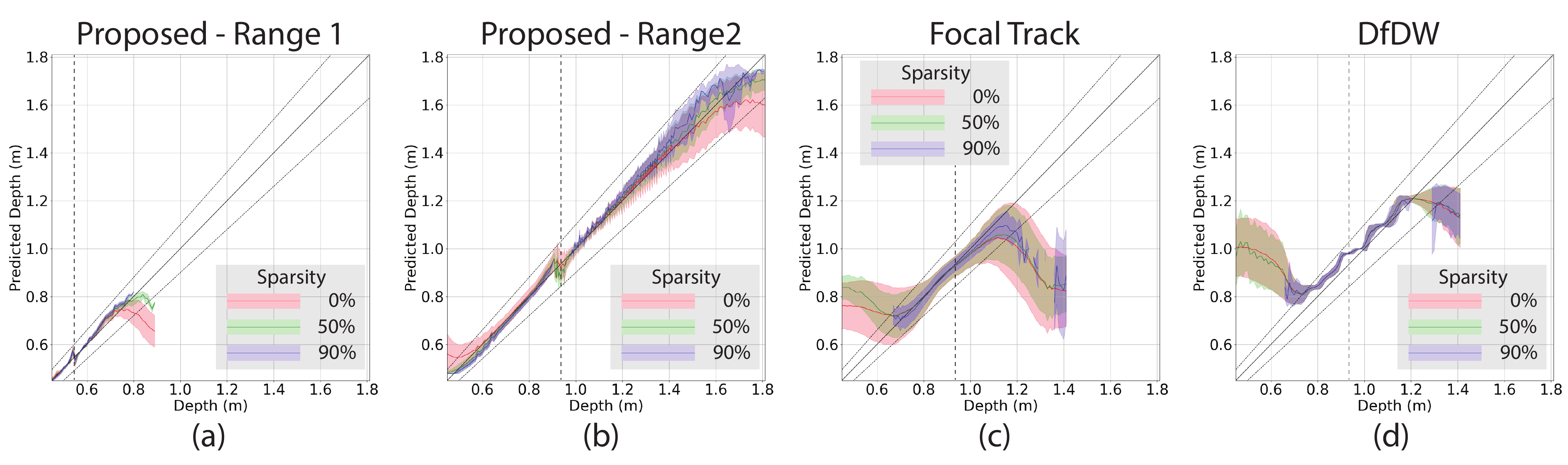}
    \vspace{-0.2in}
    \caption{Depth prediction accuracy on real data. In each plot, the solid curves indicate the mean predicted depth, and the half-widths of the color bands represent the mean deviation in prediction at each true depth. (a-b) The prototype at different offset optical powers $\rho$. The prototype can dynamically change the RoA by adjusting the offset optical power. With a closer working range, the system achieves a relatively higher depth accuracy but a smaller working range. Vice versa. Increasing the sparsity, i.e., the confidence level, elongates the working range in real data. (c-d) Focal Track~\citep{guo2017focal} and DfDW~\citep{tang2017depth} at the same offset optical power as (b), each method having its own confidence metric. We tune the parameters of these two methods to use the same receptive field dimension as ours. Comparing (b) with (c-d), the proposed system achieves more than 2x longer working range while only costing $6\%$ and $1\%$ computation of Focal Track and DfDW, respectively.}
    \label{fig:MAE_methods}
\end{figure*}

First, we quantitatively measure the prototype's depth accuracy and working range using real data. We collect images of 11 front-parallel textured planes placed at a series of known depths, whose textures are randomly sampled from a natural texture dataset~\citep{dana1999reflectance}. The system predicts a depth map from the four captured images $I(\rho+\Delta\rho, \mathnormal{A}), I(\rho-\Delta\rho, \mathnormal{A}), I(\rho, \mathnormal{A}+\Delta \mathnormal{A}), I(\rho, \mathnormal{A}-\Delta \mathnormal{A})$, where we set the finite difference steps $\Delta \rho = 0.06\;\text{dpt}$ and $\Delta \mathnormal{A} = 1\;\text{mm}$ based on the optimization result in Sec.~\ref{secsecsec:opt-diff-step}, and offset aperture radius $\mathnormal{A} = 0.25\;\text{mm}$. The offset optical power, $\rho$, can be dynamically adjusted to vary the region of accuracy (RoA). Fig.~\ref{fig:MAE_methods}a-b demonstrates the distribution of predicted depths for two offset optical powers $\rho = 10.7\;\text{dpt}$ and $10.1\;\text{dpt}$, respectively, which shows distinct regions of accuracy (RoAs). For each figure, we plot the mean of all depth predictions corresponding to the same actual depth value, $\Bar{Z}$, as the solid curves, and the mean deviation of the depth predictions for each actual depth, $\overline{|Z - \Bar{Z}|}$, as the half-width of the color band surrounding the curve. We overlay this visualization with several different confidence levels to highlight the effect of the confidence metric. Fig.~\ref{fig:MAE_methods}a-b both show a clear increase in working range when increasing the confidence level, which is consistent with the simulation analysis in Sec.~\ref{secsec:sim-conf}.

\begin{figure*}
    \includegraphics[width=\linewidth]{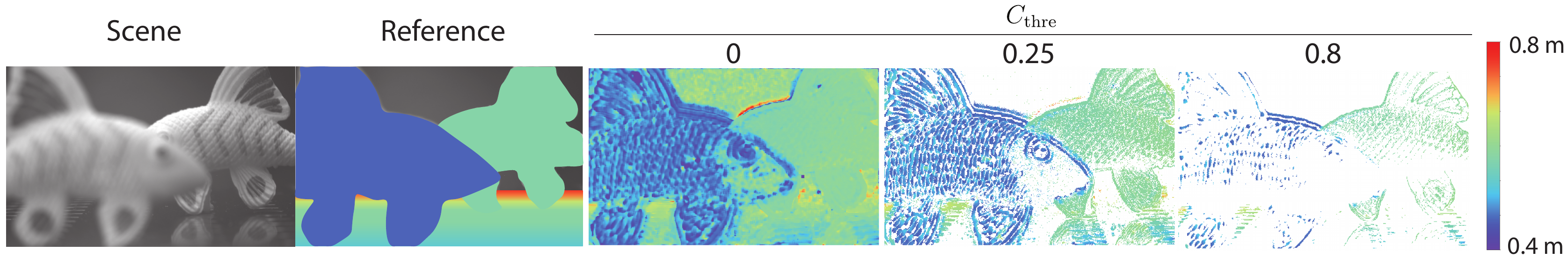}
    \caption{Depth maps of a real scene at the different confidence levels. The confidence thresholds correspond to $0\%, 50\%, 90\%$ sparsities of the depth map, respectively.}
    \label{fig:DepthMaps_sparcity}
\end{figure*}

\begin{figure*}[htbp]
    \hspace{-0.3in}
    \includegraphics[width=1.1\linewidth]{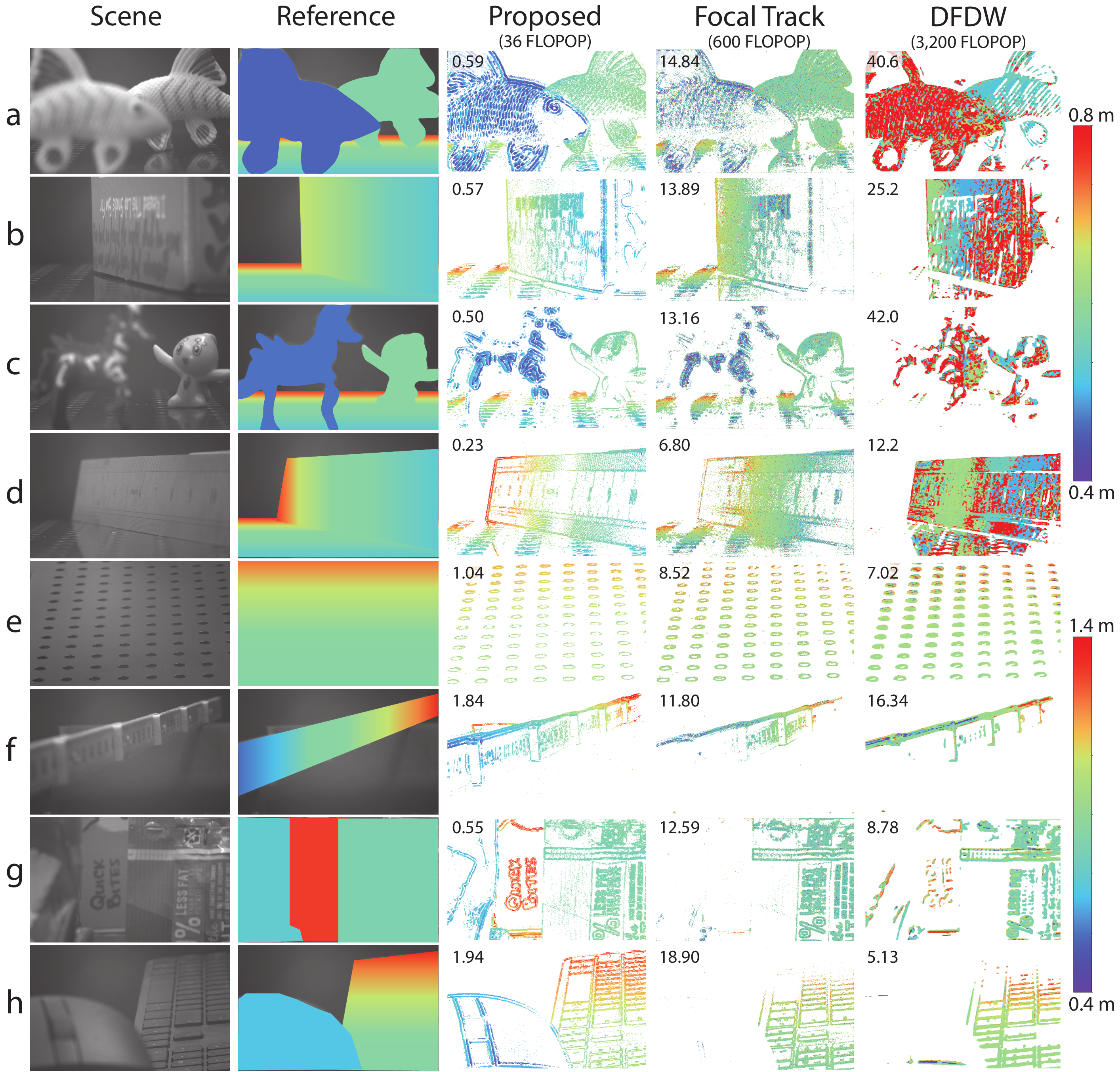}
    \caption{Depth maps of real scenes. A reference depth map for each scene estimated from manual measurement is provided in the second column. We compare the proposed method, Focal Track~\citep{guo2017focal}, and DfDW~\citep{tang2017depth} under two different working ranges, corresponding to offset optical power $\rho = 10.7\;\text{dpt}$ (a-d) and $10.1\;\text{dpt}$ (e-h). All methods use the same optical parameters and receptive field for each scene. Each depth map is filtered by the method's confidence metric. We set a constant confidence threshold for each method, $C_{\text{thre}} = 0.25, 0.7, 2500$ for ours, Focal Track, and DfDW so that the sparsity of each method's depth map is similar. The abnormal predictions of DfDW (red pixels) are due to the PSF being larger than the receptive field. The number listed in each depth map is the MAE (cm) of the confident depth predictions compared to the reference depth map. The proposed method consistently generates the most accurate depth maps while costing considerably less computation than the other two. }
    \label{fig:DepthMaps}
\end{figure*}

Furthermore, we compare the prototype's depth sensing accuracy with that of Focal Track~\citep{guo2017focal} and DfDW~\citep{tang2017depth}. Both methods only require two images with different optical powers, $I(\rho+\Delta\rho, \mathnormal{A})$ and $I(\rho-\Delta\rho, \mathnormal{A})$. We use our prototype to capture these two images with the same optical parameters,  $\rho$, $\Delta\rho$, and $\mathnormal{A}$, as in Fig.~\ref{fig:MAE_methods}b. We also tune the algorithmic parameters of Focal Track and DfDW to adopt the same receptive field dimension as the proposed method. The comparison in Fig.~\ref{fig:MAE_methods}b-d shows the proposed method has a two-time increase in the working range while maintaining a significant advantage in computational efficiency. 





Then, we qualitatively analyze the depth map generated by the proposed method. Fig.~\ref{fig:DepthMaps_sparcity} shows the effect of confidence in a typical scene captured by the prototype. As discussed in Sec.~\ref{secsec:fail}, the proposed method makes inaccurate depth predictions at textureless regions, which can be witnessed in the background of this scene. Fortunately, the confidence effectively filters out these inaccurate predictions. 

Fig.~\ref{fig:DepthMaps} visually compares a series of depth maps output by the proposed method, Focal Track~\citep{guo2017focal}, and DfDW~\citep{tang2017depth}. We test the methods with various real-world objects of different textures at two working ranges. For each scene, the three methods share the same optical parameters, $\rho$, $\Delta\rho$, and $\mathnormal{A}$, and receptive field size. For each method, we leverage its confidence metric to filter the depth map with a constant confidence threshold across all scenes. The confidence threshold for each method is determined so that different methods' depth maps are of similar sparsity. The proposed method clearly demonstrates a longer working range and the most accurate depth map despite using much fewer computational operations.




\section{Conclusion}

We present a new depth-sensing mechanism, depth from coupled optical differentiation, and a prototype sensor based on it, demonstrating unprecedented data processing efficiency and significant improvement of the working range compared to the state-of-the-art DfD methods. Limitations of the current prototype system include that the current optics require capturing four sequential images to generate a depth and confidence map, which could cause alignment issues for dynamic objects, and the depth map is sparse in areas with limited textures. Potential future work includes developing new optical systems that implement depth from coupled optical differentiation in a single shot or developing computationally efficient depth map densification algorithms.

\backmatter







\section*{Declarations}


\begin{itemize}
\item Funding: No funds, grants, or other support was received.
\item Conflict of interest/Competing interests: The authors have no relevant financial or non-financial interests to disclose.
\item Consent for publication: Not applicable.
\item Data availability: Raw images, depth maps, and confidence maps are available at \url{https://github.com/guo-research-group/cod}.
\item Materials availability: Not applicable.
\item Code availability: All code used in this paper are available at \url{https://github.com/guo-research-group/cod}.
\item Author contribution: Junjie Luo, Emma Alexander, and Qi Guo contributed to the depth from coupled optical differentiation theory development. System integration, calibration, and experimentation were performed by Junjie Luo, Yuxuan Liu, and Qi Guo. All authors contribute to the first draft of the manuscript. All authors read and approved the final manuscript.
\end{itemize}

\bibliography{refs}

\end{document}